\documentclass[conference]{IEEEtran}
\IEEEoverridecommandlockouts
\usepackage{cite}
\usepackage{amsmath,amssymb,amsfonts}
\usepackage{algorithmic}
\usepackage{graphicx}
\usepackage{textcomp}
\usepackage{xcolor}
\usepackage{booktabs}
\def\BibTeX{{\rm B\kern-.05em{\sc i\kern-.025em b}\kern-.08em
    T\kern-.1667em\lower.7ex\hbox{E}\kern-.125emX}}
\begin{document}

\title{KPNDepth: Depth Estimation of Lane Images under Complex Rainy Environment
}

\author{\IEEEauthorblockN{Zhengxu Shi}
\IEEEauthorblockA{\textit{School of Mathematical Sciences UESTC} \\
\textit{University of Electronic Science and Technology of China}\\
Chengdu, China \\
zhengxu-shi@std.uestc.edu.cn}
\and
\IEEEauthorblockN{Xiwei Liu\IEEEauthorrefmark{1}}
\IEEEauthorblockA{\textit{State Key Laboratory of Multimodal Artificial Intelligence Systems} \\
\textit{Institute of Automation, Chinese Academy of Sciences}\\
Beijing, China}
\IEEEauthorblockA{\textit{} xiwei.liu@ia.ac.cn\thanks{Corresponding Author:
{Xiwei Liu}. Email:{xiwei.liu@ia.ac.cn}}
}
}

\maketitle

\begin{abstract}
Recent advancements in deep neural networks have improved depth estimation in clear, daytime driving scenarios. However, existing methods struggle with rainy conditions due to rain streaks and fog, which distort depth estimation. This paper introduces a novel dual-layer convolutional kernel prediction network for lane depth estimation in rainy environments. It predicts two sets of kernels to mitigate depth loss and rain streak artifacts. To address the scarcity of real rainy lane data, an image synthesis algorithm, RCFLane, is presented, creating a synthetic dataset called RainKITTI. Experiments show the framework's effectiveness in complex rainy conditions.
\end{abstract}


\section{Introduction}
Depth estimation is crucial in computer vision for determining the distance of each pixel to the camera, which is key to understanding 3D scene geometry and essential for applications like autonomous driving and intelligent transportation systems. Traditional approaches, such as SLAM [1], have relied on expensive high-precision sensors like depth cameras and LiDAR for self-localization and environmental mapping. However, advances in deep neural networks have made it possible to perform dense depth estimation from a single image using a regular camera [2]. Various frameworks for monocular depth estimation have emerged, including supervised learning with 3D sensor data [3], and self-supervised and unsupervised methods leveraging binocular geometry [4] and monocular sequence pose estimation [5]. These methods are gaining popularity due to their cost-effectiveness and versatility across different scenarios\\
\indent However, monocular depth estimation faces significant challenges in complex weather conditions, such as rain streaks and local fog effects [6], which can lead to errors and loss of depth information in images. The widely used KITTI dataset [7] for monocular depth estimation contains only clear daytime road images and lacks data for rainy days and nights. Consequently, existing studies [8][9] depend on synthetic data for training. However, due to the large differences in weather, lighting, etc. between daytime lane images and rainy or night lane images, when the existing monocular depth estimation model is applied to complex environment images, there may be significant domain deviations, resulting in a notable reduction in depth estimation performance, as indicated in [14]. Therefore, existing monocular estimation research focuses on how to build a domain adaptation framework to transfer models trained on daytime images to nighttime or rainy day data sets. At the same time, unsupervised approaches that use geometric constraints instead of ground truth encounter difficulties with augmented datasets, as they are predicated on the assumption of photometric consistency [10], which is often violated in rainy conditions due to the presence of multiple light sources like streetlights, car lights, and reflective rainwater surfaces. This violation results in significant errors in photometric loss calculation and severely impacts the estimation of depth information in images. Moreover, existing rainy day depth estimation solutions still regard depth estimation in complex environments as an independent task, which makes the depth model increasingly complex.\\
\indent Therefore, stemming from the aforementioned challenges, we aim to propose a new research approach to reduce the strong dependence of depth estimation research in complex rainy environments on photometric consistency.\\
\indent When introducing complex rainy lane images, we depart from mainstream methods [11][12] that integrate synthetic and original images for self-supervised depth estimation. We segment the task into image reconstruction and depth estimation. Our efficient and precise image reconstruction network, DLKPN, trained offline, is designed to refine synthetic images before they are processed by the depth estimation network, thereby lessening the KPNDepth framework's reliance on photometric consistency and static scene assumptions.

\begin{itemize}
\item We introduce RCFLane, an algorithm accounting for environmental darkening and local fog effects in rain, to synthesize complex rainy lane images. Utilizing this, we created the RainKITTI dataset [7], which we intend to release.
\item Leveraging convolutional kernel prediction [13], we developed DLKPN to counteract rain streak artifacts in image reconstruction on RainKITTI, achieving effective image registration.
\item Our model, tested on RainKITTI, outperforms existing methods [14][15] in image reconstruction. Moreover, the KPNDepth framework demonstrates superior accuracy in depth estimation compared to baseline models [16].
\end{itemize}

\section{PROPOSED METHOD}
\subsection{RCFLane: Building Real Lane Data in Rainy Cloudy and Foggy Days}
\begin{figure}[htbp]
\centerline{\includegraphics[width=\linewidth]{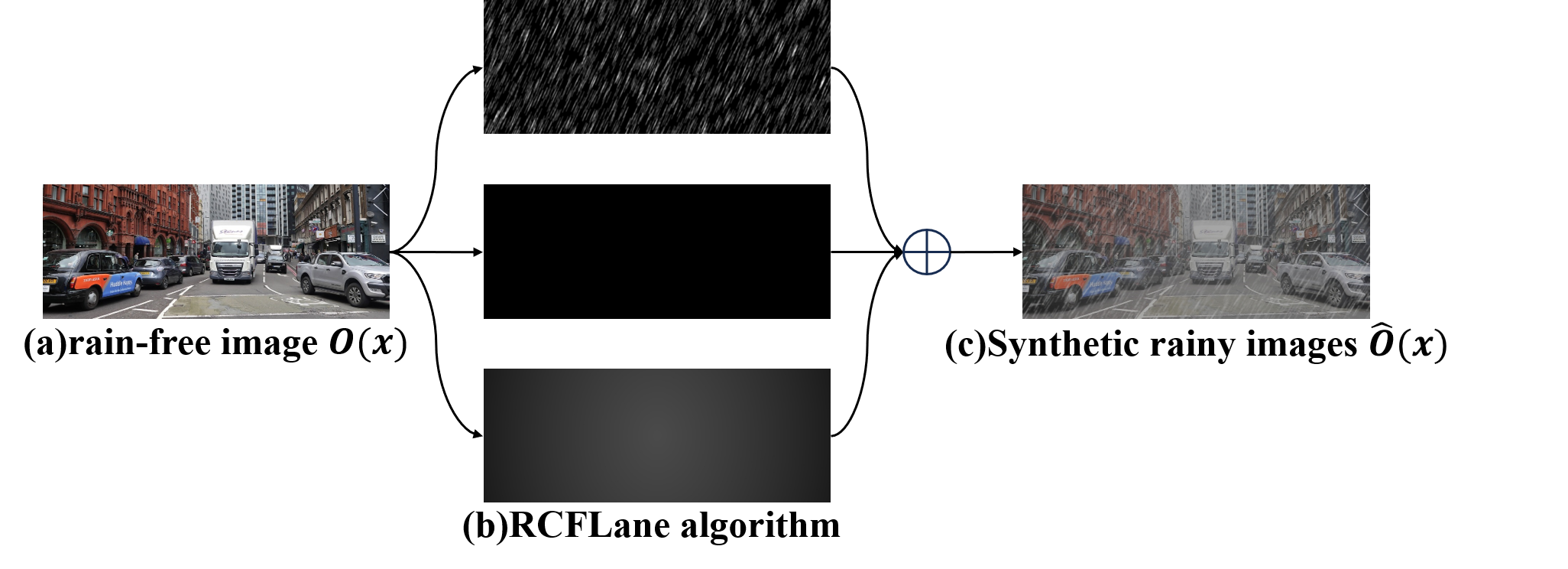}}
\caption{RCFLane Algorithm Implementation Example.}
\label{fig.1}
\end{figure}

Rain's impact on lighting and camera exposure precludes capturing realistic photos with mere variations in rain streak presence. While existing datasets typically add 2D streaks, they overlook environmental darkening and fog, essential for depth estimation tasks. The RainCityscapes dataset[17], though fog-inclusive, targets urban landscapes, not lane data, diverging from our model's training objectives. Consequently, this paper adopts the KITTI dataset to simulate realistic rain effects for constructing lane depth estimation data.\\
\indent In general, factors such as rain and fog can be considered a form of degradation, and it is reasonable to use image filtering methods to address them. Specifically, for the original lane images $O(x) \in R^{I \times J}$, based on existing research, we propose a more efficient rainy day image synthesis strategy called RCFLane:
\begin{equation}
O_1(x) = \alpha O(x) + \beta R(x)
\label{eq.1}
\end{equation}
\begin{equation}
O_2(x) = \gamma O_1(x) + (1-\gamma) D(x)
\label{eq.2}
\end{equation}
\begin{equation}
\hat{O}(x) = O_2(x) \cdot td(x) + A(1 - td(x))
\label{eq.3}
\end{equation}
 where $O_1(x)$ and $O_2(x)$ represent intermediate images in the synthetic data generation process, $\hat{O}(x)$ represents the final synthesized rainy image, $x$ denotes the two-dimensional pixel coordinates, $R(x)$ stands for the rain layer, $D(x)$ signifies the mask layer, $tr(x)$ represents the transmission layer, and $A$ represents the atmospheric light, $\alpha$, $\beta$ and $\gamma$ are used to control the retention weights of the original image, the introduction weight of the rain layer, and the introduction weight of the mask layer, respectively. For specific details, please refer to Fig. \ref{fig.1} for the detailed process.

\subsubsection{\textbf{Rain Layer}}
 In RCFLane's rain simulation, an affine transformation matrix is constructed using Gaussian blur with parameters for rotation center, angle, and scale, enabling image rain effects. Notably, the Gaussian-generated rain layer has a black backdrop, necessitating the overlay of white rain streaks, which is achieved by adjusting parameter $\alpha$ in Eq.(\ref{eq.1}) 
\begin{equation}
\alpha = \frac{255 - O(x)}{255}
\end{equation}
where $O(x)$ contains RGB pixel information across three channels. For the black parts of the rain layer, when $\alpha = 1$ all original image information is retained. Another parameter $\beta$ is manually determined and controls the introduction weight of the rain layer.
\subsubsection{\textbf{Mask Layer}}
In reality, the impact of rain on images is not only manifested in the obscuration caused by rain streaks but also in the deterioration of ambient lighting conditions. To reflect the changes in lighting conditions due to rainy days, this paper introduces a mask layer on top of the previous $O_1(x)$ in Eq.(\ref{eq.2}). The mask layer is weighted by a human-set parameter $\gamma$, to mimic the transition from day to night lighting, enhancing the realism of synthetic data for rainy conditions.
\subsubsection{\textbf{Transmission Layer}}
\begin{figure}[htbp]
\centerline{\includegraphics[width=\linewidth]{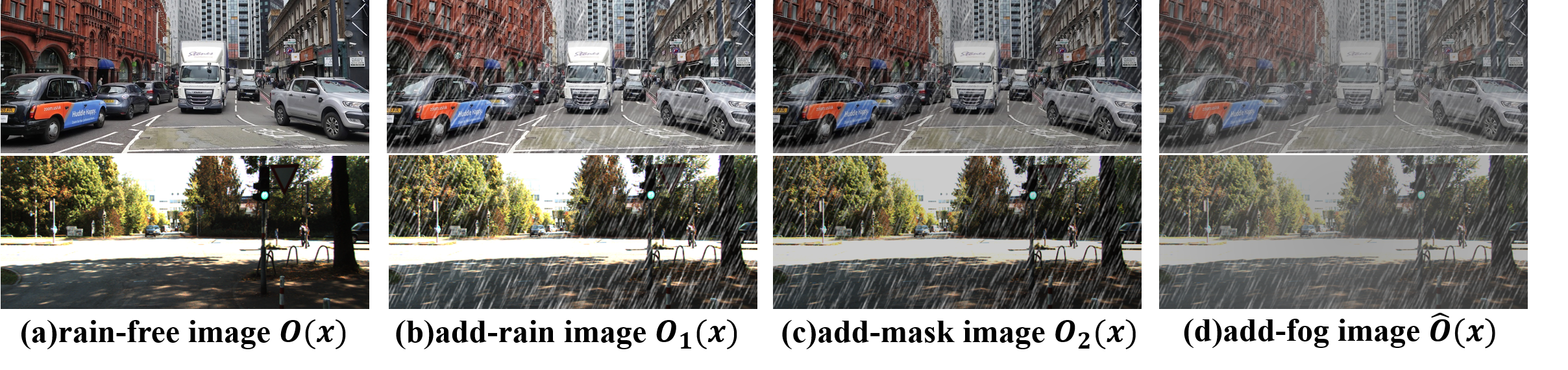}}
\caption{Example of Intermediate Image Generated by RCFLane Algorithm.}
\label{fig.2}
\end{figure}

Rain-induced fog visually imparts a characteristic to images that intensifies with increasing distance from the camera. The paper's fog addition in Eq.(\ref{eq.3}) adheres to the optical model, where light transmission is attenuated by reduced reflected energy, and atmospheric light scattering boosts brightness in fog. Therefore, the pixel value at a certain point in the foggy image consists of two parts
\begin{equation}
td(x) = e^{-\lambda d(x)}
\end{equation}
where $\lambda$ is the attenuation coefficient to control the thickness of the fog, and here $d(x)$
\begin{equation}
d(x) = - |x - x_{mid}| + S
\end{equation}
\indent The parameter $S$, inversely proportional to the distance $d(x)$ from the image center, represents fog scale. The RCFLane algorithm employs a central fogging point for rainy lane image synthesis, diverging from depth-guided methods like those in [6]. This choice is motivated by two factors: it circumvents the need for accurate depth data, enhancing flexibility, and it aligns with the center's significance in lane imagery, representing the farthest point from the camera. Given the distance-dependent nature of rain-induced fog, the algorithm rationally assumes peak fog density at the center, tapering off at the periphery, thus justifying the focus on central fog.\\
\indent Fig. \ref{fig.2} shows the intermediate image obtained by the above three steps.

\subsection{Dual-layer Pixel-wise Convolution Kernel Prediction Network}
Rain and the resulting illumination reduction and local fog are forms of image degradation we aim to restore before depth estimation to mitigate rain-induced depth increases and artifacts. The need for real-time depth prediction in modern tasks also demands efficient inference times for our restoration algorithm.\\
\indent Drawing from recent advances in image restoration and rain removal [13][18][19], we propose a kernel prediction network that predicts convolutional kernels rather than direct pixel values for the target pixels. The approach, supported by [13][18][19], is advantageous for faster convergence and inference, meeting our study's goals.\\
\indent The KPN network operates on each pixel $p$ to be denoised in the synthesized rainy image $\hat{O}(x) \in R^{I \times J}$
\begin{equation}
\tilde{O}(p) = W(p) \ast K_{p}
\end{equation}
where $\tilde{O}(p)$ represents the pixel value at point $p$ in the reconstructed image, $W(p) \in R^{K \times K}$  represents a $K \times K$ window centered at $p$ in $\hat{O}(x)$, $\ast$ denotes pixel-wise filtering operation, and represents the pixel-wise convolutional kernel predicted for pixel $p$.\\
\begin{figure}[htbp]
\centerline{\includegraphics[width=\linewidth]{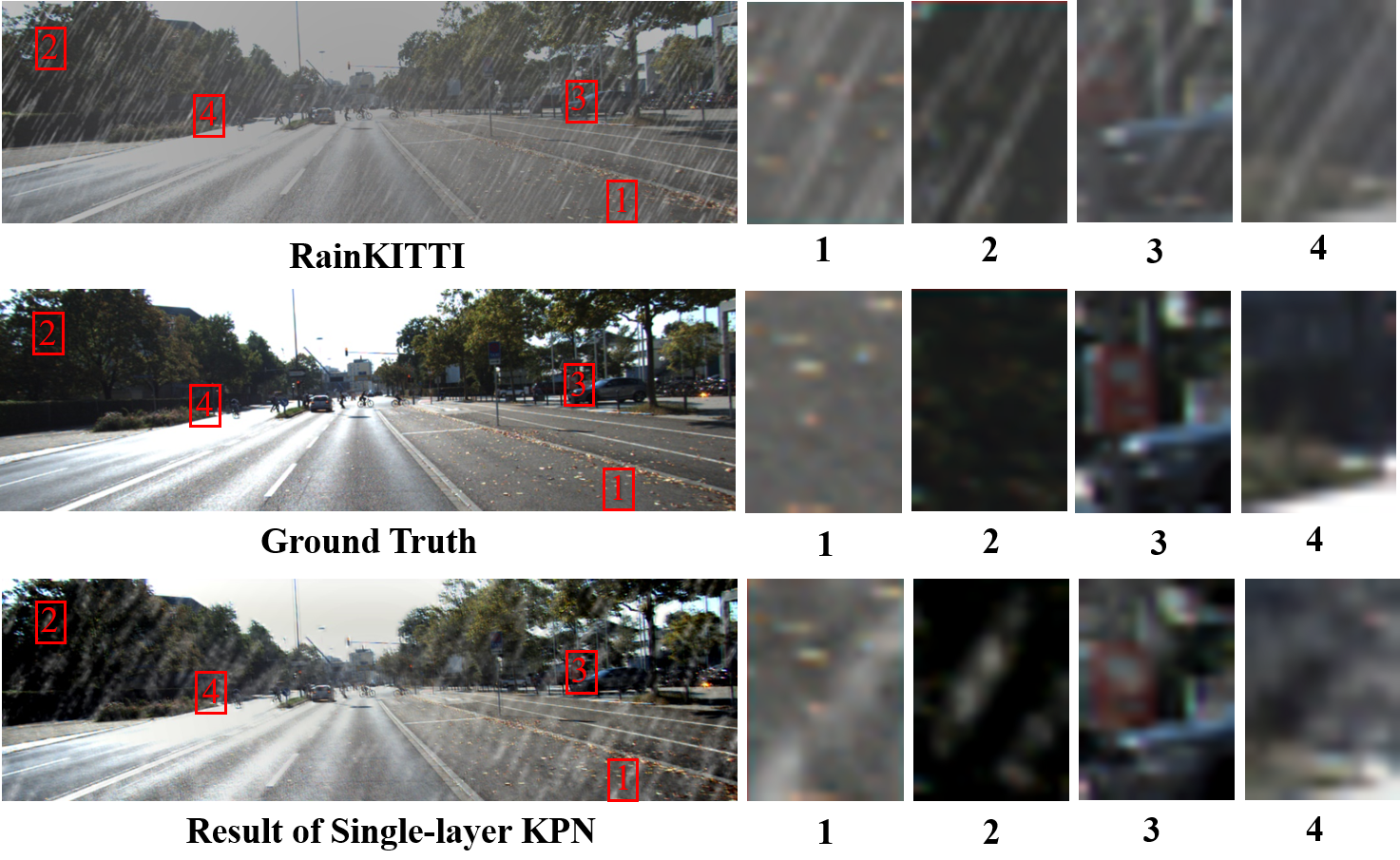}}
\caption{Display of Prediction Results of Single-layer KPN on RainKITTI.}
\label{fig.3}
\end{figure}\\
\begin{figure*}[htbp]
\centerline{\includegraphics[width=\linewidth]{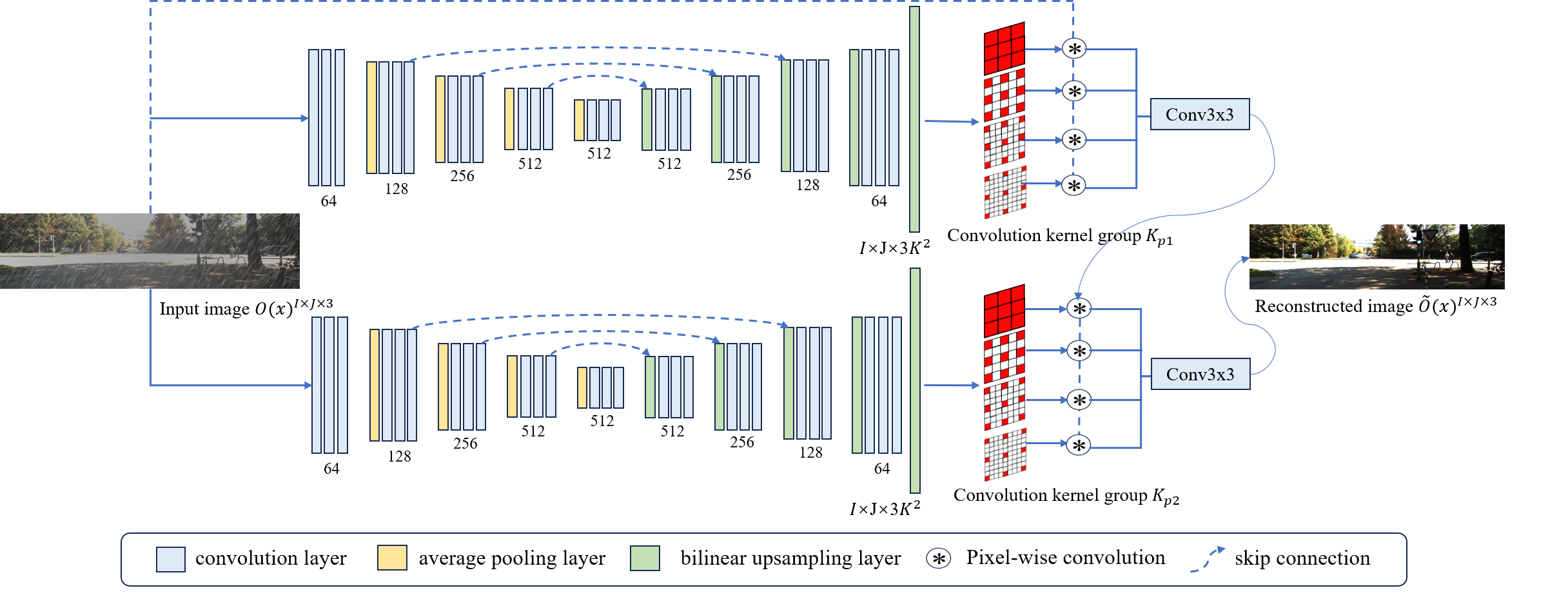}}
\caption{DLKPN Structure Diagram.}
\label{fig.4}
\end{figure*}
\indent Experiments revealed that degradation in $\hat{O}(x)$ relative to $O(x)$ is not just from rain streaks, and a single KPN network with pixel-wise kernels is insufficient for complete degradation explanation, leaving residual artifacts. Fig. \ref{fig.3} illustrates this on RainKITTI data, with full results in Section IV. To address this, we developed the Dual-layer Pixel-wise Convolution Kernel Prediction Network (DLKPN), trained on distinct datasets to recover degradation and remove rain streak artifacts. For pixel $p$ in $\hat{O}(x)$, DLKPN predicts two independent kernel sets $K_{p1}$ and $K_{p2}$:
\begin{equation}
\tilde{O}_{m}(p) = W(p) \ast K_{p1}
\end{equation}
\begin{equation}
\tilde{O}_{f}(p) = W(\tilde{O}_{m}(p)) \ast K_{p2}
\end{equation}
where $\tilde{O}_{m}(p)$ and $\tilde{O}_{f}(p)$ represent the predicted values of pixel $p$ in the first and second layers of the network, respectively. In practical computation, we adopt the hierarchical dilation strategy from [18]. We expand the pixel-wise convolution kernels $K_{p1}$ and $K_{p2}$ using a coefficient $r = 0, 1, 2, 3$ to construct dilated convolution kernels, which are aimed at filtering out rain streaks at multiple scales. Fig. \ref{fig.4} illustrates the specific structure of DLKPN.

\subsection{Added Depth Estimation Model of DLKPN}
\begin{figure*}[htbp]
\centerline{\includegraphics[width=\linewidth]{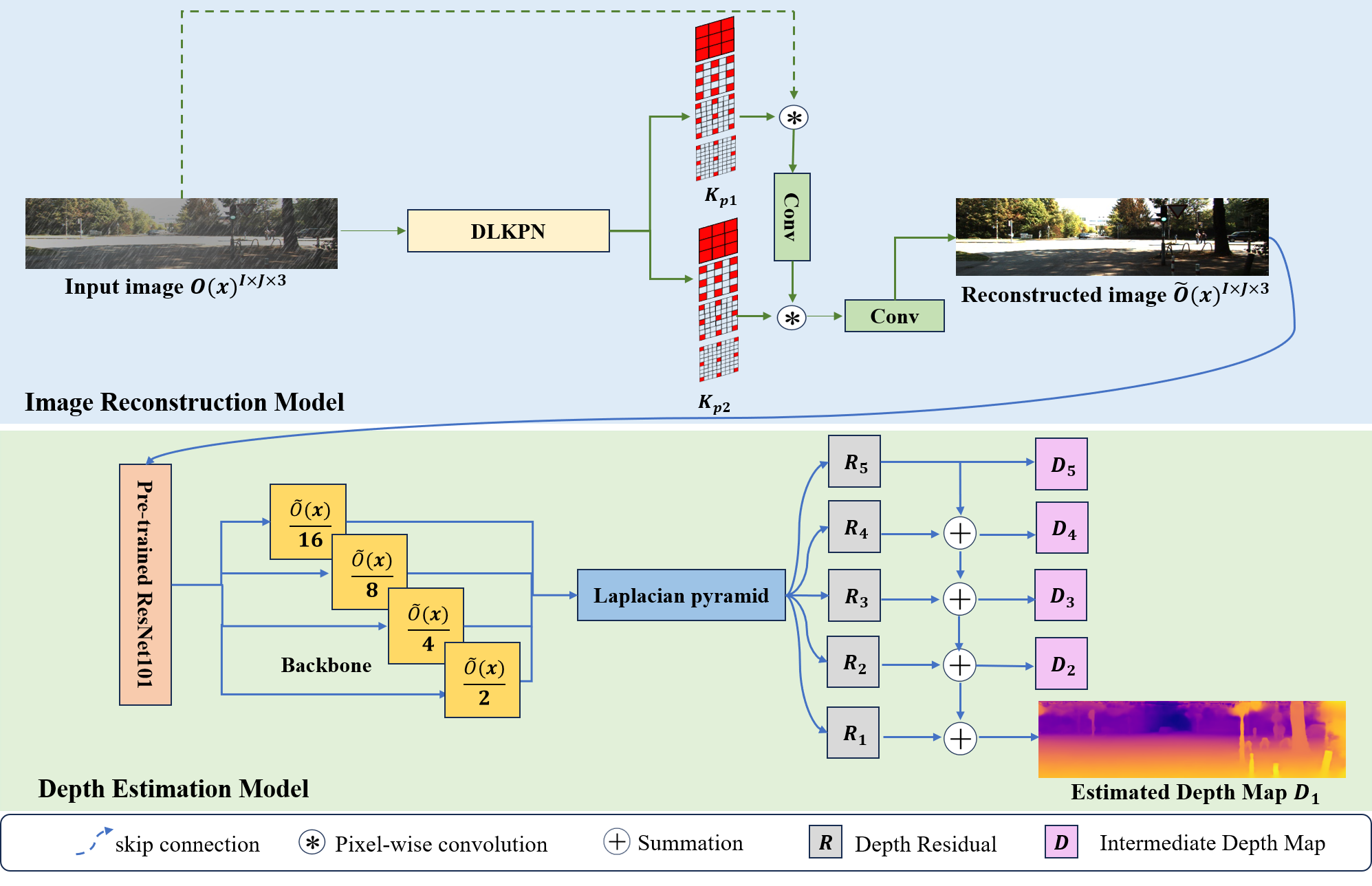}}
\caption{The Overall Structure of The KPNDepth Framework.}
\label{fig.5}
\end{figure*}

The DLKPN model, optimized for rapid inference, is proposed for rainy lane image reconstruction to meet real-time depth estimation demands. We will train the DLKPN model offline using the RainKITTI dataset and combine it with conventional models to improve rainy-day depth estimation accuracy, termed as KPNDepth. We chose LapDepth[16] as the baseline, utilizing Laplacian pyramids to decode depth residuals from multi-stream features, refining the depth image progressively. Refer to Fig. \ref{fig.5} for the KPNDepth framework's architecture.

\section{EXPERIMENTS}
\subsection{Datasets}
This paper tackles the challenge of rainy lane depth estimation, an area where specialized datasets are limited. To address this, we leveraged the KITTI dataset [7], a well-established benchmark for autonomous driving vision algorithms that includes diverse urban, rural, and highway scenes. Utilizing the RCFLane algorithm, we synthesized the RainKITTI dataset from the $2011_09_28$ and $2011_09_29$ sequences, carefully selecting 820 images while preserving their original resolution of 1238 x 374 pixels. This dataset was then split into training and test sets, with the ground truth depth information retained for thorough model evaluation.\\
\subsection{Experimental Setup}
All our experiments and inferences were conducted on a single NVIDIA GeForce RTX 4090 GPU using the PyTorch framework. The network architecture of the depth estimation model remains consistent with LapDepth [16].\\
\subsection{Metrics}
\textbf{Rainy Day Image Reconstruction Metrics}: We use peak signal-to-noise ratio (PSNR) and structural similarity (SSIM), which are commonly used in the field of image reconstruction, as quantitative evaluation metrics for all datasets. Here, the larger the PSNR and SSIM are, the better the rain removal effect is. Times (ms) represents the processing time of a single image.\\
\indent \textbf{Rainy Day Lane Depth Estimation Metrics}: For the quantitative evaluation of the depth estimation effect, we use six indicators introduced by Eigen et al. [20] \textbf{Abs Rel}, \textbf{Sq Rel}, \textbf{RMSE}, \textbf{RMSE log}, \textbf{Accuracy} and \textbf{log10}, which have been widely used and verified in the field of depth estimation.

\subsection{Results}

\begin{table}[h]
\caption{Comparison with Current Image Reconstruction Models for Complex Rainy Weather Images}
\centering
\begin{tabular}{ccccl}
\toprule
\ & \multicolumn{2}{c}{\textbf{Quantitative Evaluation Metric}} & \\
\cmidrule(lr){2-3}\cmidrule(lr){1-1} \cmidrule(lr){4-4}
 Method& SSIM & PSNR & Times (ms)\\
\midrule
DRSformer[14] & 0.708 & 22.381 & 754.72 \\
PreNet[15] & 0.685 & 16.491 & 95.35 \\
PReNet\_r[15] & 0.680 & 16.537 & 94.35 \\
PRN[15] & 0.684 & 16.440 & 33.05 \\
PRN\_r[15] & 0.681 & 16.396 & \textbf{32.85} \\
DLKPN & \textbf{0.941} & \textbf{30.601} & 152.89 \\
\bottomrule
\end{tabular}
\label{table:1}
\end{table}
\begin{table*}[h]
\caption{Comparison with Depth Estimation Baseline Model in Complex Rainy Conditions}
\centering
\begin{tabular}{ccccccccl}
\toprule
\ & \multicolumn{4}{c}{\textbf{Error Metrics (Lower is better)}} &\multicolumn{3}{c}{\textbf{Accuracy (Higher is better)}} \\
\cmidrule(lr){2-5}\cmidrule(lr){1-1} \cmidrule(lr){6-8}
 Method& Abs Rel & Sq Rel & RMSE & RMSE log&$\delta < 1.25^1$&$\delta < 1.25^2$&$\delta < 1.25^3$\\
\midrule
LapDepth[16] & 0.172 & 1.383 & 6.304&0.298&0.708&0.846&0.921 \\
KPNDepth(S) & 0.110 & 0.715 & 4.276&0.202&0.838&0.930&0.981 \\
KPNDepth(D) & \textbf{0.066} &\textbf{ 0.447} & \textbf{2.897}&\textbf{0.136}&\textbf{0.941}&\textbf{0.977}&\textbf{0.988}\\
\bottomrule
\end{tabular}
\label{table:2}
\end{table*}
\begin{figure*}[htbp]
\centerline{\includegraphics[width=0.8\linewidth]{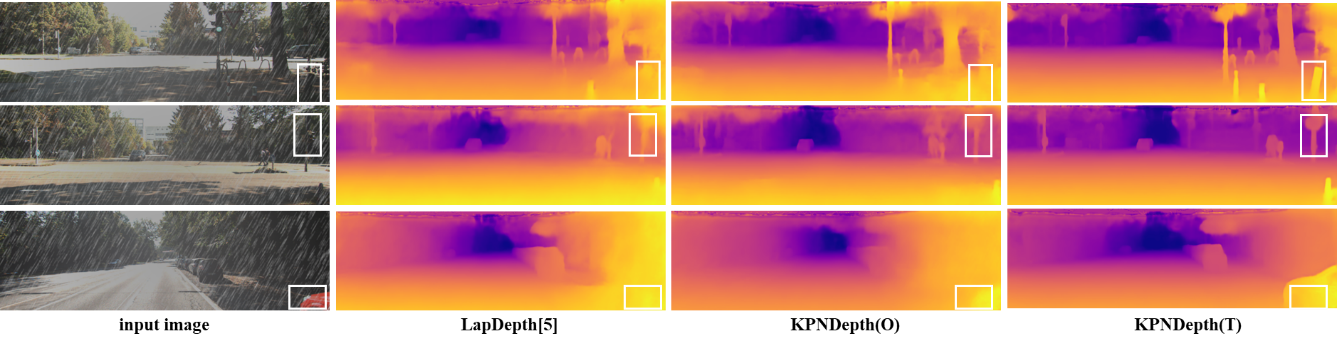}}
\caption{Comparison of Depth Maps Output by Depth Estimation Models on RainKITTI.}
\label{fig.6}
\end{figure*}

\subsubsection{\textbf{Image Reconstruction on RainKITTI}}
\ 
\newline 
\indent The experimental results, presented in Table \ref{table:1}, demonstrate the performance of the DLKPN method on the RainKITTI dataset, which lacks comparable prior data due to its novel construction. The absence of open-sourced studies, such as [17], on rain curtain fog effects further limits direct comparisons. To assess DLKPN's efficacy, we benchmarked it against the SOTA model DRSformer[14] and the lightweight PreNet[15] on both Quantitative Evaluation Metrics and inference latency.\\
\indent The table includes variations of PreNet[15], which differ by the presence of recursive layers. Inference latency, measured in Times(ms), indicates the processing time for a single image. DLKPN outperformed [14] in SSIM and PSNR by accounting for rain-induced environmental darkening and fog, evident in the visual quality of restored images. Despite a higher latency than the lightweight PRN\_r, DLKPN's 152.89ms single-image inference time is 4.94 times faster than DRSformer, making it a balanced choice given its superior SSIM and PSNR.

\subsubsection{\textbf{Depth Estimation of Lane in Rainy Days}}
\ 
\newline 
\indent The relevant experimental results are shown in Table \ref{table:2}, , and the depth estimation results are displayed in Fig. \ref{fig.6}. Due to the scarcity of related works on depth estimation in complex environments, there is a lack of existing models and results for comparison. Inspired by similar work [21], we will focus on analyzing the improvement of KPNDepth compared to LapDepth, which serves as the baseline model.\\
\indent In Table \ref{table:2} KPNDepth(S) indicates the depth estimation with only a single-layer pre-trained KPN introduced, while KPNDepth(D) represents using pre-trained DLKPN for prior image reconstruction during depth estimation. According to the results provided in the table, we observe that the LapDepth network with the introduction of pre-trained DLKPN achieves optimal results in both model loss and accuracy. Furthermore, from the depth prediction image in Fig \ref{fig.6} it is evident that the object edges predicted by KPNDepth(D) are clearer and more detailed. During the experiments, we found that due to the influence of rain curtains, the overall depth estimation values of rainy images tend to be erroneously inflated. By comparing the predicted results in Fig. \ref{fig.6}, we notice that KPNDepth(D) effectively avoids this issue among the three models.

\begin{table}[h]
\caption{Results of a Quantitative Study of Ablation on RainKITTI}
\centering
\begin{tabular}{ccccl}
\toprule
 Method& SSIM & PSNR & Times (ms)\\
\midrule
EfDeRain[18] & 0.699 & 21.999 & \textbf{68.54} \\
DLKPN & \textbf{0.941} &\textbf{ 30.601} & 152.89 \\
\bottomrule
\end{tabular}
\label{table:3}
\end{table}

\subsubsection{\textbf{Ablation Study}}
\ 
\newline 
\indent In developing the DLKPN network, we adopted the strategy of kernel prediction over direct pixel value prediction, as per [13][18][19]. Expanding on [18]'s single KPN model, we added an independent set of kernels to eliminate rain streak artifacts.\\
\indent To assess the contribution of the new kernels, we conducted ablation studies, with results in Table \ref{table:3}. These show that DLKPN's dual kernels notably improve reconstruction quality and robustness on lane images in adverse weather.\\
\indent The DLKPN model, while outperforming the baseline in computational accuracy due to its enhanced convolutional operations, incurs longer computation times for single image processing. In scenarios with simple lighting, this extra computational load is deemed unnecessary. Given the DLKPN framework's dual-module structure, we propose a preliminary assessment of the lane environment before inference. For environments with minimal rain streaks and fog, computation is restricted to the single-layer KPN network, reducing unnecessary latency and improving model efficiency in real-world applications.

\section{CONCLUSION}
In this paper, we address the issue of lane image depth estimation in highly complex rainy environments, which is of great significance for expanding the application scope of depth estimation models and enhancing their robustness in complex lane scenarios.\\
\indent Our study analyzes the limitations of existing datasets and introduces the RainKITTI dataset synthesized using the RCFane algorithm to emulate real complex rainy conditions. We propose a novel pre-trained DLKPN network for rainy image reconstruction, which, when combined with conventional depth estimation models, forms the KPNDepth framework. This framework employs pixel-wise convolution to recover depth information obscured by rain streaks and enhance image reconstruction in rainy lane scenarios. It demonstrates improved performance over current methods [14][15] and surpasses the baseline model [16] in depth estimation accuracy.\\
\indent Furthermore, the DLKPN framework presented in this study, trained on offline data, demonstrates superior inference latency and strong transferability, offering a viable solution for depth estimation in rainy conditions. It has been evaluated on standard rainy datasets like Rain100, yielding promising results. However, it has limitations, notably the absence of night-time rainy driving scenarios. The complex light reflections in these low-light conditions can significantly disrupt image reconstruction and depth estimation, presenting a challenge for accurately recovering depth information in such environments.

\section*{Acknowledgment}
This work was partially supported by the National Science and Technology Major Project (2022ZD0117102).

\end{document}